\documentclass[letterpaper, 10 pt, conference]{ieeeconf}  

\IEEEoverridecommandlockouts                              

\usepackage{xcolor}
\usepackage[table,xcdraw]{ colortbl}
\usepackage{multirow}
\usepackage{caption}

\usepackage{graphicx} 
\usepackage{epsfig} 
\usepackage{amsmath} 
\usepackage{amssymb}  
\usepackage{multirow}
\usepackage{todonotes}
\usepackage{dirtytalk}
\graphicspath{ {./fig/} }
\usepackage{hyperref}
\hypersetup{
    colorlinks=true,
}

\title{\LARGE \bf
AirTrack: Onboard Deep Learning Framework for Long-Range Aircraft Detection and Tracking}

\author{Sourish Ghosh$^{1}$, Jay Patrikar$^{1}$, Brady Moon$^{1}$, Milad Moghassem Hamidi$^{1}$, and Sebastian Scherer$^{1}$
\thanks{*This work was supported by the Army Research Laboratory (ARL) grant: \textbf{ARL IDIQ-HA W911QX-20-F0106}. This material is based upon work supported by the National Science Foundation Graduate Research Fellowship under Grant No. DGE1745016.}
\thanks{$^{1}$The authors are with the Robotics Institute, Carnegie Mellon University, Pittsburgh, PA, USA 15213.
        {\tt\small \{sourishg, jaypat, bradym, mmoghass, basti\}@andrew.cmu.edu}}%
}

\begin{document}

\maketitle
\thispagestyle{empty}
\pagestyle{empty}

\begin{abstract}

Detect-and-Avoid (DAA) capabilities are critical for safe operations of unmanned aircraft systems (UAS). This paper introduces, AirTrack, a real-time vision-only detect and tracking framework that respects the size, weight, and power (SWaP) constraints of sUAS systems. Given the low Signal-to-Noise ratios (SNR) of far away aircraft, we propose using full resolution images in a deep learning framework that aligns successive images to remove ego-motion. The aligned images are then used downstream in cascaded primary and secondary classifiers to improve detection and tracking performance on multiple metrics. We show that AirTrack outperforms state-of-the art baselines on the Amazon Airborne Object Tracking (AOT) Dataset. Multiple real world flight tests with a Cessna 182 interacting with general aviation traffic and additional near-collision flight tests with a Bell helicopter flying towards a UAS in a controlled setting showcase that the proposed approach satisfies the newly introduced ASTM F3442/F3442M standard for DAA. Empirical evaluations show that our system has a probability of track of more than 95\% up to a range of 700m. \\
\href{https://youtu.be/H3lL_Wjxjpw}{[Video]}\footnote{ \href{https://youtu.be/H3lL_Wjxjpw}{Video: https://youtu.be/bMw5nUGL5GQ}}

\end{abstract}

\section{Introduction}

Mid-air collision (MAC) and near mid-air collision (NMAC) risk are concerns for both manned and unmanned aircraft operations, especially in low-altitude airspace. 
Detect-and-Avoid (DAA), also commonly referred to as \textit{sense} and \textit{avoid}, is defined as \say{the capability of an aircraft to remain \textit{well} \textit{clear} from and avoid collisions with other airborne traffic.}~\cite{saa} The well clear boundary as defined by NASA~\cite{munoz2014family} mathematically characterizes a volume, referred to as the Well Clear Violation, such that if aircraft pairs jointly occupy this volume, they are considered to be in a Well Clear Violation. In visual flight rules conditions, NMAC/MAC threat mitigation is carried out by a pilot visually \textit{detecting} and \textit{avoiding} other aircraft to remain \textit{well clear}~\cite{munoz2015daidalus} of them. Typically for medium to large airborne systems, an active onboard collision avoidance system such as the Traffic Alert and Collision Avoidance System or the Airborne Collision Avoidance System is used and relies on transponders installed in cooperative aircraft. However, not all airborne threats can be tracked using transponders. Rogue drones, balloons, light aircraft, and inoperative transponders present a threat to reliable operations. This makes DAA an essential requirement especially for beyond visual line of sight operations in the National Airspace System. 

Human vision is the last line of defence against a mid-air collision and is thus critical for aviation safety. Therefore, in order to assist pilots in mitigating mid-air collision threats, machine vision can be used to alert pilots of potential aircraft and objects in the sky. Due to size, weight, and power (SWaP) constraints of UASs, radar is often not a feasible solution. Machine vision has been a promising direction for research in this domain~\cite{mcfadyen2016survey} based on success of CNN-based networks.
\begin{figure}[!t]
    \centering
    \includegraphics[width=0.95\columnwidth]{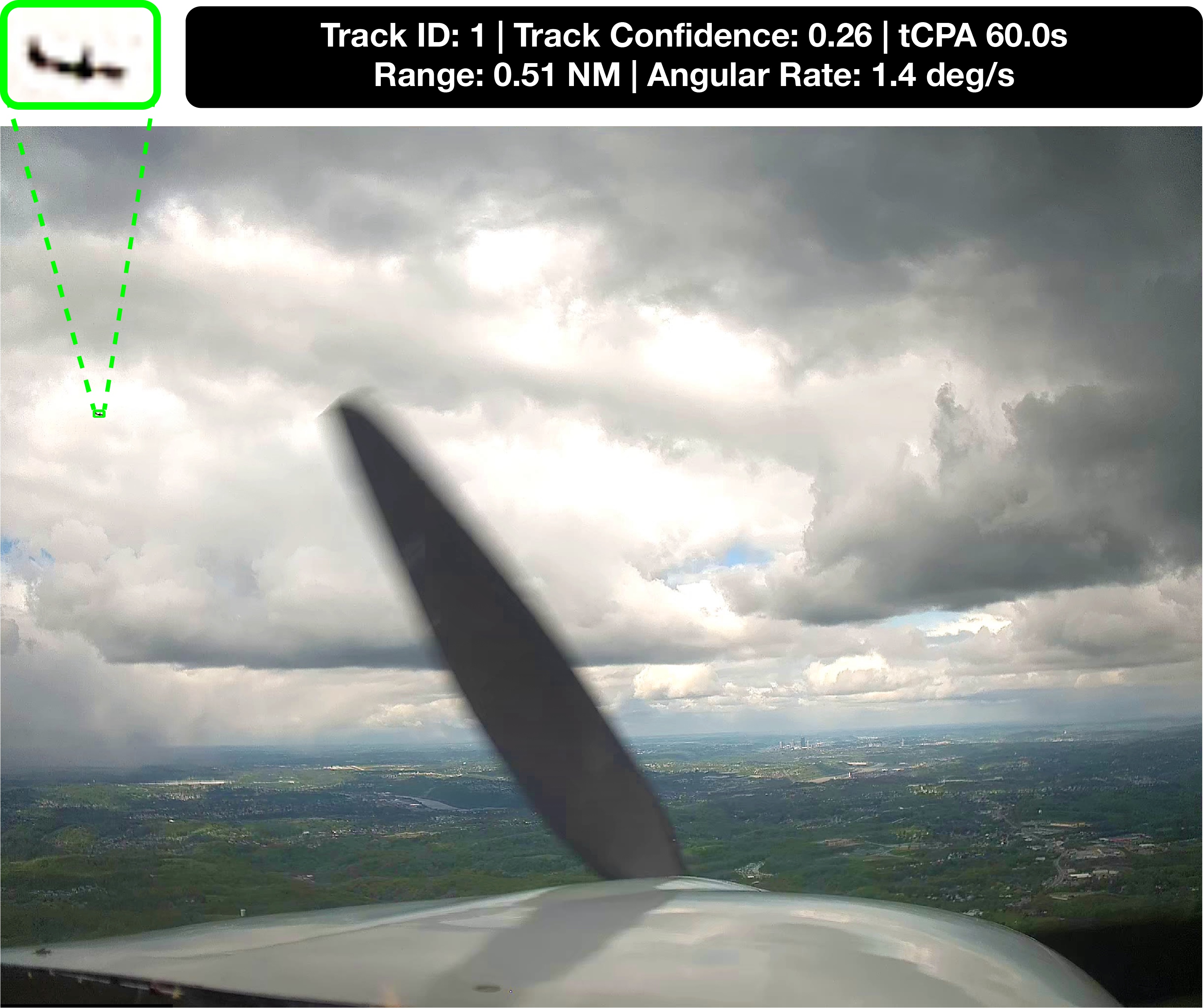}
    \caption{\small Snapshot showing visual detection and tracking of an intruder in the real world tests with a Cessna 182 aircraft using our system. The bounding box width is less than 15px in an image of resolution $2448\times 2048$ highlights the challenges of visual DAA. Top left corner shows the zoomed in crop around the detected object. The information box shows the relevant DAA parameters computed by the AirTrack framework.}
    \label{fig:cover_pic}
\end{figure}

\begin{figure*}[t]
    \centering
	\includegraphics[clip, trim=0.0cm 0cm 0.0cm 0cm, width=1.00\textwidth]{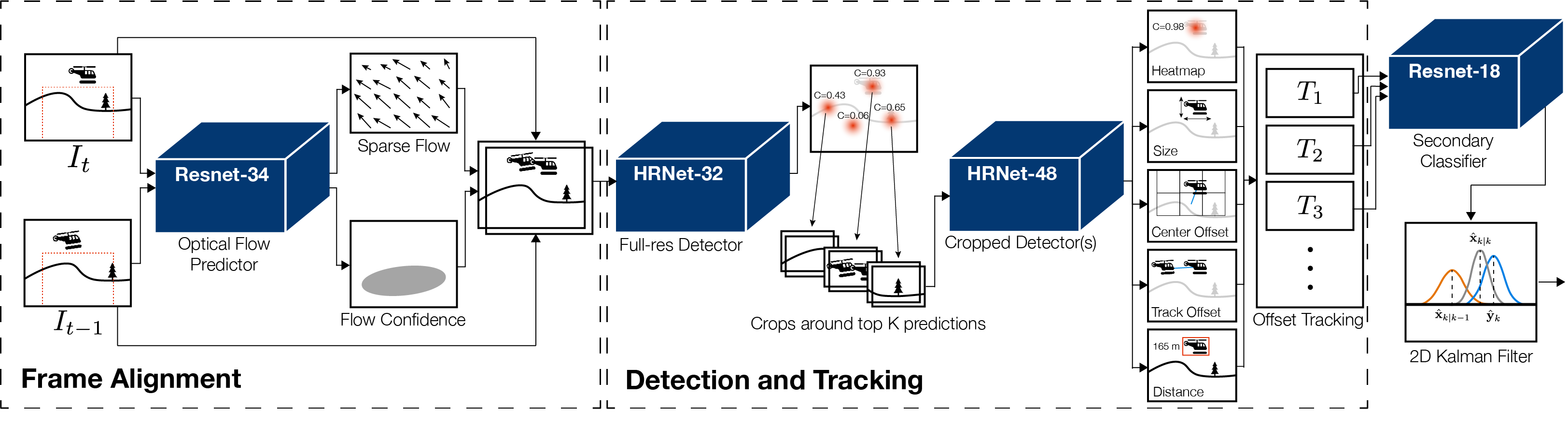}
	\caption{\small AirTrack architecture showing the various internal components. The architecture uses frame alignments, cascaded detection modules and a secondary classifier. Finally a 2D Kalman filter is used to get the relevant intruder DAA parameters. Shaded gray boxes are the modules are the white boxes are the input/output variables. \label{fig:system_design}}
	\vspace{-0.5em}
\end{figure*}

In order to standardize DAA capabilities, performance requirements F3442/F3442M - 20 have been published by ASTM~\cite{f3442} which define safe DAA operations of UAS as having a maximum dimension of less than or equal to $25$ft, operating at airspeeds below $100$kts, and of any configuration or category. This standard does not define a specific DAA architecture. According to this standard, for safe DAA operations at a particular intruder range, the probability of track must be greater than 95\% and the range estimation error of the intruder must be within 15\%, along with a maximum cap on the angular rate error based on ownship specifications.

In this work, we present AirTrack, a machine vision-based solution for detecting and tracking aircraft that aims to satisfy the ASTM performance standards. We propose an end-to-end deep learning solution that leverages recent advances in machine vision to enable a real-time SWaP-C solution for DAA. Given the low Signal-to-Noise ratios (SNR) of the detection, we propose using full resolution images in a deep learning framework that aligns successive images to remove ego-motion. The aligned images are then used downstream in cascaded primary and secondary classifiers to improve detection performance on multiple metrics. The key contribution of this work is three-fold: (1) an end-to-end vision-based aircraft detection and tracking system that uses cascaded detection modules, (2) a system of primary and secondary classification modules that maintain high precision in novel scenarios via false-positive mining and retraining, and (3) real-world flight tests with onboard SWaP hardware on both a general aviation Cessna 182 performing standard general aviation flight behaviors as well as on a UAS performing dedicated near-collision experiments with a Bell helicopter intruder flying towards the UAS in controlled settings, with both showcasing satisfaction of ASTM F3442/F3442M standards for DAA. 
\section{Related Work}

Vision-based DAA has a rich literature background spanning classical pipe-line based techniques to modern era end-to-end deep learning techniques.

\subsection{Classical Approaches}

Traditionally, the main solutions for visual DAA included a modular design with various well-known techniques commonly used in classical computer vision~\cite{lai2013characterization, dey2010passive, fasano2014morphological, carnie2006image, lai2011airborne, mejias2010vision}. The primary modules are the following in sequential order: frame stabilization, background-foreground estimation using morphological operations, temporal filtering, detection, and tracking. The frame stabilization step is typically handled using either optical flow-based approaches~\cite{mccandless1999detection} or image registration using feature matching~\cite{reilly2010detection, schubert2014robust}.
~\cite{rozantsev2015flying} used regression-based motion compensation both in the horizontal and vertical directions, and the morphological operations are used to enhance the signal of the intruder aircraft. ~\cite{lai2013characterization, dey2010passive, carnie2006image} used background subtraction. Machine learning based approaches have been also used to learn descriptors of the aircraft~\cite{rozantsev2015flying, dey2010passive, petridis2008learning} using methods such as SVMs. Naively using morphological operations would result in false positives, hence a temporal filtering stage is used. Track-before-detect is a commonly used technique for detecting low-SNR targets in infrared imagery~\cite{fernandez1990detecting} and several have tried using it in vision-based DAA. Approaches for tracking include using Kalman filter banks~\cite{nussberger2014aerial}, Hidden Markov Models~\cite{lai2013characterization, lai2008hidden, molloy2017detection}, and Viterbi-based filtering~\cite{carnie2006image, lai2013characterization}.

\subsection{Deep Learning Approaches}
\subsubsection{Detection}
Modern visual DAA approaches rely on deep neural networks based object detection and tracking methods. Detection is typically accomplished using convolution neural networks (CNN)s~\cite{chen2014aircraft, hwang2018aircraft, stojnic2021method, james2018learning}. The key challenge is visually detecting small objects in large resolution images. Due to the very low SNR, standard anchor-based methods such as YOLO and R-CNN are not ideal for small object detection. Keypoint-based architectures~\cite{duan2019centernet} are much more suitable for small objects. Related work can also be found in the domain of face detection~\cite{hu2017finding}, aerial imagery~\cite{liang2021learning}, and pedestrian tracking~\cite{zhu2019feature}. Since computational efficiency is a key requirement for DAA systems, state of the art approaches include fully convolution networks with heatmap prediction~\cite{james2018learning, james2019below}. 

\subsubsection{Tracking}
Tracking-by-detection is the commonly used paradigm for aircraft tracking~\cite{bewley2016simple}. A tracking management system is typically maintained to account for birth and death of new tracks, and associate new detections to existing tracks using the Hungarian algorithm~\cite{lai2013characterization, dey2010passive}. In multi-object tracking, typically a metric like Intersection-over-Union (IoU) for bounding boxes is used to determine the assignment, but for small objects the IoU metric becomes too sensitive to slight deviations in bounding box positions. In that case, an integrated detection and tracking approach such as~\cite{zhou2020tracking} is a better choice, which our approach is based on.

This manuscript is organised as follows: Section II details prior approaches in visual DAA including classical approaches along with modern state-of-the-art deep learning-based approaches. Section III explains the proposed approach in detail, and Section IV includes the specifics to our implementation. Section V contains an evaluation of the proposed method based on relevant metrics and real-world testing. Finally, section VI presents concluding remarks and future work.

\section{Methodology}
 The overall system design of AirTrack is shown in Fig.~\ref{fig:system_design} and consists of the following four sequential modules: (1) Frame Alignment, (2) Detection and Tracking, (3) Secondary Classification, and (4) Intruder State Update. Let the inputs be two successive grayscale image frames $I_{t}, I_{t-1}\in\mathbb{R}^{H\times W}$ where $H\times W$ are the dimensions of the input frames. We utilize the full image resolution during inference to maximize the chances of detection at long ranges ($\geq$ 1km). The final outputs of the system are a list of tracked objects with bounding box coordinates, track ID, 2D Kalman Filter state (containing pixel-level velocity and acceleration), estimated range, and time to closest point of approach (tCPA). The following subsections describe each of the modules in detail.

\subsection{Frame Alignment}

In order to help distiquish the foreground objects from the background, one must align successive frames in a video so that the ego-motion of the camera can be discarded. This is done with the help of a frame alignment module that predicts the optical flow between two successive image frames and the confidence of the predicted flow. 

This module takes as input the current and previous input frames $I_{t}, I_{t-1}\in\mathbb{R}^{H\times W}$, where $H$ and $W$ are the input image height and width respectively. Since the sky is mostly textureless, it does not provide much addition information for computing the background flow. The input images are thus cropped from the center-bottom covering most of the high-texture details present below the horizon using a fixed-size crop of size 2048 × 1280. The backbone architecture for alignment is a ResNet-34 with two prediction heads: (1) optical flow offsets $\mathbf{F}_t\in\mathbb{R}^{2\times H\times W}$, (2) confidence heatmap of offsets $\mathbf{C}_t\in\mathbb{R}^{H\times W}$. The prediction is made at 1/32 scale of the input, and low confidence offset predictions are rejected. 

For training, 75\% of the input image tuples are created by data-augmentation. The data-augmentation involves generating a random affine homography by sampling the parameters from a normal distribution. This random homography is used to warp an image frame and thus create a training sample. The remaining 25\% of the input image tuples are picked as successive frames from the dataset. The target homography for the neural network is generated by first computing the Lucas-Kanade optical flow (OpenCV) and then finding the affine transform. The training objective minimized is the following:
\begin{equation*}
    \frac{1}{N}\sum\left(\mathbf{C}_t \circ \text{MSE}\left(\mathbf{F}_t, \hat{\mathbf{F}}_t\right)\right)
\end{equation*}
where $\hat{\mathbf{F}}_t$ and $\mathbf{F}_t$ are the predicted and ground truth optical flow respectively, and MSE denotes the mean squared error. 

\subsection{Detection}
The detection architecture is made up of two cascaded detection modules. The primary detection module takes in two full resolution aligned image frames $H \times W$, and the secondary detection module takes as input a smaller crop of $h\times w$, $h, w\approx H/5, W/5$ around the top $k$ detector outputs (based on confidence) of the primary module. The cascaded network is fully-convolutional, and the output scale is 1/8 of the input resolution. It outputs five maps: (1) center heatmap encoding the center of the object, (2) bounding box size, (3) center offset from grid to center of object, (4) track offset of object center from the previous frame, (5) distance of the object in log scale. Each of these five heads is trained with a separate loss function. 

The target center heatmaps during training are rendered using a Gaussian kernel. For distant objects, this results in a single-pixel render which is not helpful for training. Therefore a minimum box size of $3\times 3$ is used for center rendering. The center heatmap is trained using the focal loss~\cite{lin2017focal} for handling large class imbalances:
\begin{equation*}
    L_h = \frac{1}{N}\sum_{xy}\left\{\begin{array}{lr}
        (1 - \hat{H}_{xy})^\alpha\log \hat{H}_{xy},  \text{if }  H_{xy}=1\\
        (1 - H_{xy})^\beta\hat{H}_{xy}^\alpha\log (1 - \hat{H}_{xy}),  \text{otherwise}
        \end{array}\right\}
\end{equation*}
where $x,y$ are the pixel locations in ground-truth and predicted heatmaps $H, \hat{H}$, and $\alpha, \beta$ are the focal loss parameters. The peaks of this predicted heatmap are the object centers, and we choose the corresponding values from the other predicted heads based on the pixel locations of the peaks. 

The bounding box size prediction is regressed by minimizing the following objective:
\begin{equation*}
    L_{\text{size}} = \frac{1}{N}\sum_{i=1}^N\left| \hat{s}_{\mathbf{p}_i} - \mathbf{s}_i \right|
\end{equation*}
where $\hat{s}_{\mathbf{p}_i}$ is the predicted box size at pixel location $\mathbf{p}_i$ and $\mathbf{s}_i$ is the ground truth box size. L1 error is also minimized for the center offset prediction:
\begin{equation*}
    L_{\text{off}} = \frac{1}{N}\sum_{i=1}^N\left| \hat{o}_{\mathbf{p}_i} - o_{\mathbf{p}_i} \right|
\end{equation*}
where $\hat{o}_{\mathbf{p}_i}$ is the predicted offset and $o_{\mathbf{p}_i}$ is the ground truth offset. The track offset loss is again an L1 loss:
\begin{equation*}
    L_{\text{track}} = \frac{1}{N}\sum_{i=1}^N\left| \hat{T}_{\mathbf{p}_i^{(t)}} - \left(\mathbf{p}_i^{(t-1)} - \mathbf{p}_i^{(t)}\right) \right|
\end{equation*}
where $\hat{T}_{\mathbf{p}_i^{(t)}}$ is the predicted track offset and $\mathbf{p}_i^{(t-1)} - \mathbf{p}_i^{(t)}$ denotes the change in the location of the center pixel from the previous frame. Finally, the distance prediction is trained by optimizing the following L2 loss:
\begin{equation*}
    L_{\text{dist}} = \frac{1}{N}\sum_{i=1}^N\left| \log\hat{D}_{\mathbf{p}_i} - \log D_{\mathbf{p}_i} \right|
\end{equation*}
where $\hat{D}_{\mathbf{p}_i}$ and $D_{\mathbf{p}_i}$ are the predicted and ground truth distance. To help with training stability, we predict log distance to scale the large distance values. The overall training objective that is minimized is thus:
\begin{equation*}
    L_{\text{total}} = w_1L_h + w_2L_{\text{size}} + w_3L_{\text{off}} + w_4L_{\text{track}} + w_5L_{\text{dist}}
\end{equation*}
where $w_i, i=\{1,\dots,5\}$ are hyper-parameters.

\subsection{Tracking}
The tracking approach builds on top of the offset tracking vector approaches \cite{zhou2020tracking}.
The key idea in this algorithm is to use the track offset vector to associate current frame detection to a list of existing tracks. The power of this algorithm lies in its simplicity. Using the predicted track offset vector, we subtract the offset from the current object center to recover the center location in the previous frame. Then we compare the previous frame center with the offset-adjusted center and check if the distance between them falls below a certain threshold $\kappa$. If it does, we propagate the existing track ID to the current detection and mark it as matched. Otherwise, we spawn a new track ID with the current detection and proceed. 

\subsection{Secondary Classifier}
A ResNet-18 module is used as a secondary classifier for false-positive rejection. The input to the network is a fixed-size crop (padded and resized) around the bounding boxes detected by the object detectors. It is a binary classification network that predicts whether the input crop is an aircraft or not. The training data for this network is collected from the results of the detectors. We mine false-positive samples to train the classifier for better false-positive rejection. We use focal loss~\cite{lin2017focal} to train the model since a training batch contains more false-positive samples than true positives. The secondary classifier aims to improve the overall precision of the system and to allow quick retraining using novel data rather than training the detector, which would take some time. This enables one to run the object detectors at a low confidence threshold to improve recall.

\subsection{Intruder State Estimation}
The final stage of our DAA module is intruder state estimation. This module maintains an internal state of each intruder detected using a 2D Kalman Filter with a constant acceleration motion model. For offset tracking, this filter is necessary to compute pixel-level velocity and acceleration. The angular-rate is computed as:
\begin{equation*}
    r = \theta\sqrt{\dot{x}^2 + \dot{y}^2}
\end{equation*}
where $\theta$ is the degree/pixels ratio of the camera, and $\dot{x}, \dot{y}$ denotes the pixel velocity of the object center. This module also computes the time to closest point of approach:
\begin{equation*}
    \text{tCPA}(i, j) = \frac{t_j - t_i}{-1 + \sqrt{\frac{a_j}{a_i}}}
\end{equation*}
where $a_k, t_k$ denotes the area of the bounding box and time at frame $k$ respectively. This tCPA formulation is based on~\cite{glozman2021vision} and the key idea is that as the bounding box area of the intruder increases in size (meaning that the object is getting closer), the tCPA value decreases. tCPA is inversely proportional to the rate of change of the square root of the bounding box area.

\section{Implementation}
Since we are dealing with small objects, the choice of the backbone architectures is very crucial for good performance. The key requirements are learning high-resolution image features that can be used to identify tiny objects against the sky and ground clutter while also learning enough context to ignore false positives. This section provides the necessary implementation details.
\subsection{Dataset}
For training, we use the Airborne Object Tracking (AOT) Dataset. This dataset was released in 2021 as part of the Airborne Object Tracking Challenge~\cite{aotchallenge} organized by AIcrowd in partnership with Amazon Prime Air. It is a collection of around 5000 flight sequences of 120 seconds each at 10Hz resulting in 164 hours of total flight data. There are a total of 3.3M+ labeled image frames containing airborne objects. The image resolution is 2448x2048, and the images are 8-bit grayscale. Along with bounding box and class labels, the annotations also include range information of the aircraft for a part of the dataset. The range mainly varies from 600 to 2000 meters (25-75 percentiles). The area of the objects labeled vary from 4 to 1000 sq pixels. Among the planned airborne encounters, among 55\% of them would qualify as potential collision trajectories. $80\%$ of the targets are above the horizon, $1\%$ on the horizon, and $19\%$ below the horizon. The dataset also captures different sky and visibility conditions; $69\%$ of the sequences have good visibility, $26\%$ have medium visibility, and 5\% exhibit poor visibility conditions. We use a train/val split of 90/10 to train and evaluate models on this dataset. 

\subsection{Training Details}
 The backbone architectures used for the primary and secondary detectors are HRNet~\cite{wang2020deep}, and it is an ideal choice for this problem because it fuses high-level and low-level parallel convolutional feature maps using a unique fusion operation that preserves high-resolution features with enough low-level context. The full-resolution detector predicts the initial heatmap. Then $512\times 512$ crops are taken around the top $K$ ($K=4$) heatmap peaks and formed into an input batch for the cropped detector. The cropped detector is typically chosen to be a heavier neural network with more parameters since it is operating on a lower-resolution image.

The frame alignment and detection neural networks are trained using the SGD optimzer~\cite{defazio2021adaptivity} with cosine annealing warm restarts~\cite{loshchilov2017sgdr} as the learning rate scheduling strategy. $512\times 512$ image crops are used to train all the detection networks. Since they are fully-convolutional, smaller random image-crop pairs are sufficient for training. The sampling strategy of the image batches is important for a good overall performance of the detector. For training batch samples, 50\% are chosen with random crops, 25\% are chosen around hard false-positives, and the remaining 25\% are crops taken around true aircraft locations. False-positive predictions during training with a confidence score over 0.2 are saved for mining and are sampled for training in later epochs to improve the precision of the model. The models were trained on a Tesla P100 GPU with 16GB of GPU memory.

\subsection{AirTrack Hardware}
\begin{figure*}[t]
    \centering
    \includegraphics[width=\textwidth]{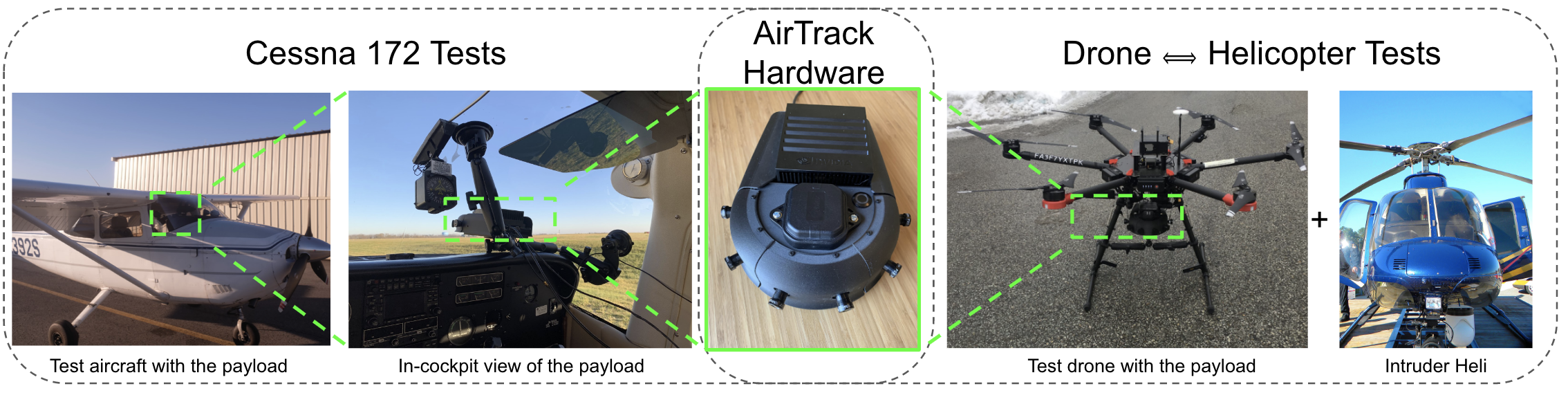}
    \caption{\small Overview of the different field tests. Flight tests were performed with onboard SWaP hardware on both a general aviation Cessna 172 performing standard general aviation flight behaviors as well as on a UAS performing dedicated near-collision experiments with a Bell 407 helicopter intruder flying towards the UAS in controlled settings, with both showcasing satisfaction of ASTM F3442/F3442M standards for DAA.  }
    \label{fig:field}
    \vspace{-0.7em}
\end{figure*}
The AirTrack algorithm is designed to run on a custom build SWaP-C hardware platform as shown in Fig. \ref{fig:field}. The platform contains six Sony IMX264 cameras spanning a total of approximately $220^\circ$ FOV horizontally and $48^\circ$ vertically. The payload uses an NVIDIA Xavier AGX 32GB dev kit to handle the image processing. The payload also contains a 3DM-GQ7 GNS/INSS module with a dual-antenna setup for state estimation and ground truth. The platform is designed to run standalone with an external power source.  

\section{Evaluation Experiments}
To showcase the efficacy of AirTrack, we present results on held-out AOT dataset as well as real world flight tests with large scale aircraft. Comparisons with established baselines, ablation studies and above/below horizon detection performances are provided. We compare out approach to two baseline methods. Siam-MOT based on Siamese multi-object tracking\cite{shuai2021siammot} and YOLOv5\cite{glenn_jocher_2020_4154370} trained on the AOT dataset. The confidence threshold for prediction was set at 0.5. Table~\ref{tab:kpi} outlines the key detection and tracking metrics over which we evaluate the methods.
\begin{table}[t]
\centering
\begin{tabular}{|l|l|l|}
\hline
\textbf{Metric} & \textbf{Description} & \textbf{Domain} \\ \hline
\textbf{P} $\uparrow$ & Precision & $[0, 1]$ \\ 
\textbf{R} $\uparrow$ & Recall & $[0, 1]$ \\ 
\textbf{EDR} $\uparrow$& Encounter Detection Rate & $[0, 1]$ \\
\textbf{FPPI} $\downarrow$& False positives per image & $\geq 0$ \\ 
\textbf{IDSPI} $\downarrow$ & Track ID switches per image & $\geq 0$ \\
\textbf{ARE} $\downarrow$ & Angular rate error & $\geq 0$ (deg/s) \\ \hline
\end{tabular}
\caption{ \small Key Performance Metrics}
\label{tab:kpi}
\vspace{-0.5em}
\end{table}
\subsection{Flight Tests}
We perform two kinds of flight tests as shown in Fig \ref{fig:field}. For the first kind, we install the AirTrack hardware on a general aviation Cessna 182. The aircraft is flown closer to airports and is allowed to interact with cooperative general aviation air traffic. The second set of tests involve a UAS flying towards an intruder helicopter. The UAS and the helicopter start at opposite ends of the runway at different altitudes. They converge and break right to ensure collision avoidance. Post-hoc, we manually label both of these for evaluation.

    

\subsection{Qualitative Comparative Results}
For qualitative evaluations we combine results from held-out AOT and real world flight tests to provide a complete analysis. For the encounter detection rate, an encounter is considered valid if the airborne object is consistently tracked for at least 3 seconds before its range falls below 2000ft. Table~\ref{tab:results} outlines the overall results of the proposed system along with the two baselines. From the table, it can be deduced that YOLOv5+SORT has the worst performance. This is because YOLOv5, which is an excellent \textit{state-of-the-art} object detector in general, is not the most ideal for detecting small objects. Our system outperforms the baselines in all the metrics mentioned earlier. Using the secondary classifier (SC) greatly improves the precision of the system while only taking a slight reduction in the recall.
\begin{table}[t]
\centering
\begin{tabular}{|c|cc|cc|}
\hline
\multirow{2}{*}{\textbf{Metric}} & \multicolumn{2}{c|}{\textbf{Baselines}}              & \multicolumn{2}{c|}{\textbf{Our System}}                   \\ \cline{2-5} 
                  & \multicolumn{1}{c|}{YOLOv5+SORT} & Siam-MOT & \multicolumn{1}{c|}{SC: off}         & SC: on          \\ \hline
\textbf{P}                 & \multicolumn{1}{c|}{0.8436}      & 0.9758   & \multicolumn{1}{c|}{0.9532}          & \textbf{0.9916} \\ \hline
\textbf{R}                 & \multicolumn{1}{c|}{0.3326}      & 0.4253   & \multicolumn{1}{c|}{\textbf{0.4776}} & 0.4634          \\ \hline
\textbf{EDR}               & \multicolumn{1}{c|}{0.8867}      & 0.938    & \multicolumn{1}{c|}{0.9542}          & \textbf{0.9639} \\ \hline
\textbf{FPPI}              & \multicolumn{1}{c|}{0.0987}      & 0.0050   & \multicolumn{1}{c|}{0.0111}          & \textbf{0.0018} \\ \hline
\textbf{IDSPI}             & \multicolumn{1}{c|}{0.0872}      & 0.0011   & \multicolumn{1}{c|}{0.0010}          & \textbf{0.0009} \\ \hline
\textbf{ARE}               & \multicolumn{1}{c|}{1.15}        & 0.89     & \multicolumn{1}{c|}{0.87}            & \textbf{0.85}   \\ \hline
\end{tabular}
\caption{ \small Comparison of proposed method w/ baselines.}
\vspace{-0.5em}
\label{tab:results}
\end{table}

\subsection{Above/Below Horizon Results}
Detecting aircraft above and below the horizon poses significantly different challenges, the latter being the more difficult due to the presence of background clutter. Table~\ref{tab:horizon} outlines the detection metrics based on \textit{above} or \textit{below} the horizon cases. We can observe that the precision is steady across both domains while the recall is much lower for below-horizon detections being a more difficult task. This is a key area to improve upon in future work.
\begin{table}[b]
\centering
\begin{tabular}{|c|cc|cc|}
\hline
\multirow{2}{*}{\textbf{}} & \multicolumn{2}{c|}{\textbf{Above Horizon}}    & \multicolumn{2}{c|}{\textbf{Below Horizon}}    \\ \cline{2-5} 
                           & \multicolumn{1}{c|}{SC: off} & SC: on & \multicolumn{1}{c|}{SC: off} & SC: on \\ \hline
\textbf{P}                          & \multicolumn{1}{c|}{0.94}    & \textbf{0.99}   & \multicolumn{1}{c|}{0.94}    & \textbf{0.99}   \\ \hline
\textbf{R}                          & \multicolumn{1}{c|}{\textbf{0.63}}    & 0.59   & \multicolumn{1}{c|}{\textbf{0.38}}    & 0.36   \\ \hline
\textbf{FPPI}                       & \multicolumn{1}{c|}{0.0122}  & \textbf{0.0012} & \multicolumn{1}{c|}{0.0175}  & \textbf{0.0026} \\ \hline
\textbf{IDSPI}                      & \multicolumn{1}{c|}{\textbf{0.0071}}  & 0.0073 & \multicolumn{1}{c|}{0.0015}  & \textbf{0.0011} \\ \hline
\end{tabular}
\caption{\label{tab:horizon}  \small Performance of the proposed system based on metrics calculated separately for above and below horizon scenarios.}
\end{table}
We can also observe the impact of the secondary classifier improving false-positive rejection by reducing the FPPI number by almost 10x. It however does reduce the recall slightly.

\subsection{ASTM Interpretation}
\begin{figure}[]
    \centering
    
   \includegraphics[clip, trim=0.0cm 0cm 0.0cm 0.0cm, width=0.45\textwidth]{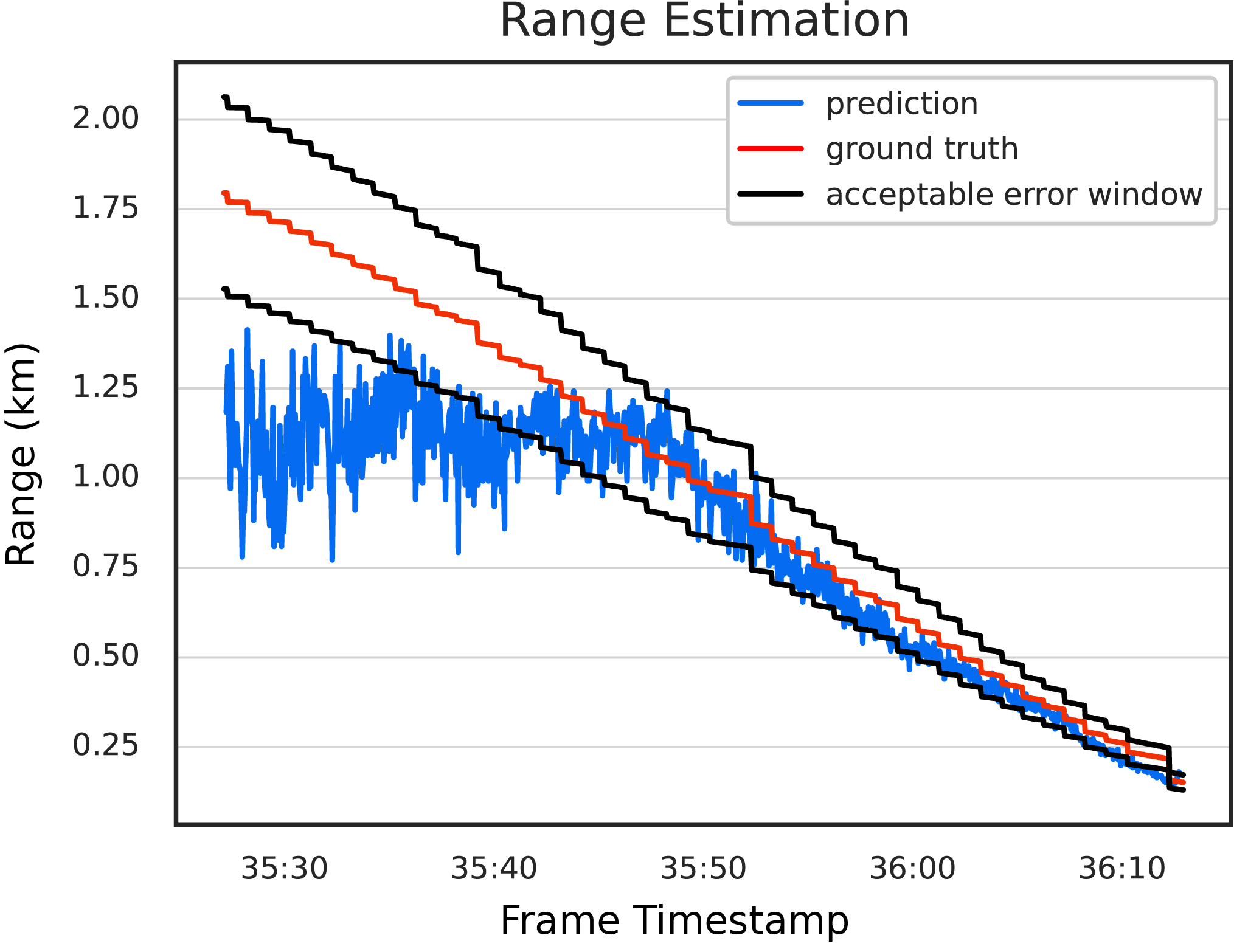}
	\caption{\small Range estimation plots compared to ground-truth. We note that the range estimation deteriorates after 1.2km typically. Black line denotes the 15\% error margin for interpretation with regards to ASTM F3442/F3442M standard\label{fig:range_estimation}}
\end{figure}

\begin{figure}[!h]
    \centering
	\includegraphics[width=0.45\textwidth]{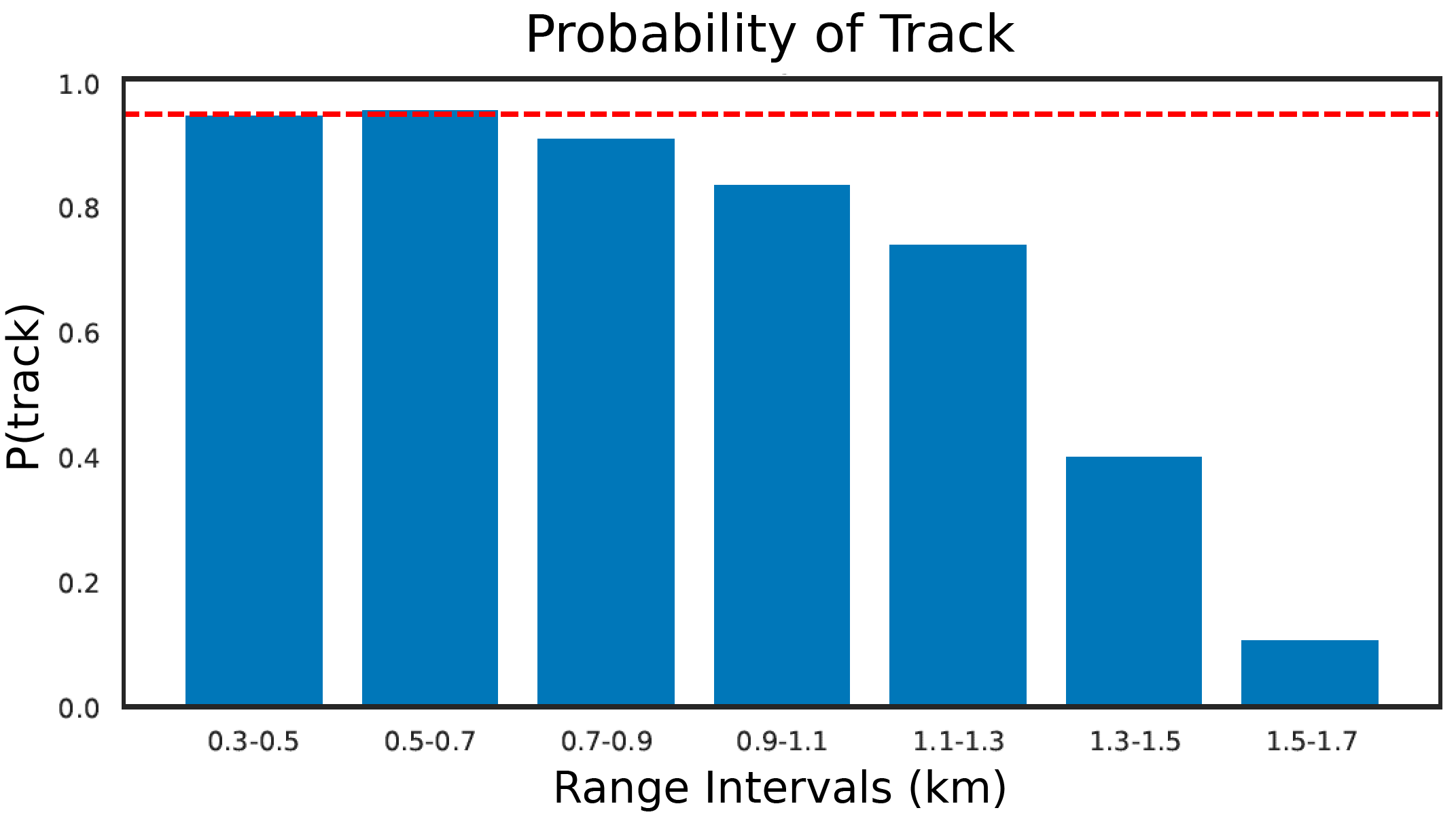}
	\caption{\small P(track) vs range. The probability of track (recall) reduces with increasing distance. Dashed red line shows the 95\% level. P(track) $\geq$ 95\% upto 700m. \label{fig:range_recall}}
 \vspace{-1.0em}
\end{figure}

We provide and interpret results from the perspective of the ASTM standards.  
 The range error is computed as a fraction of the ground truth range. For the tests we observe that the median range error obeys the allowable error threshold of 15\% up to a range of 1.5km. Fig.~\ref{fig:range_estimation} shows a qualitative range plot for detected aircraft from the real world flight tests for a single sequence.
We can observe that the range prediction quality gets worse with increasing object distance, especially with distances of over 1.25km. Fig.~\ref{fig:range_recall} shows the probability of track (recall) of the proposed system based on range intervals for the AOT dataset. The probability of track remains more than 95\% up to a range of 700m. 
    
    

Based on our quantitative results, we interpret our results with regard to the ASTM F3442/F3442M standard~\cite{f3442}. This standard was created for unmanned aircraft (UA) with a maximum dimension $\leq 25$ft, operating at airspeeds below 100kts, and of any configuration or category. 
This standard found only range estimation and angular-rate error to be the only two key performance indicators for DAA and specifies a minimum intruder tracking probability of 95\%. Although it does not specify how this probability was computed, we interpret it as the tracker recall value (see Fig.~\ref{fig:range_recall}). Based on the recall value, our system has been shown to have a probability of track of more than 95\% up to a range of 700m. We also know our system angular-rate error is within 0.9deg/s. Thus based on these two error metrics, we can use the standard to check which aircraft classes can be used for low risk-ratio operations using our DAA surveillance system. Table~\ref{tab:ownship_classes} outlines the minimum track range requirement for different autonomous UAS specs based on a maximum angular-rate error of 0.9deg/s.
\begin{table}[t]
\begin{tabular}{|ccc|c|}
\hline
\multicolumn{3}{|c|}{\textbf{Ownship Specifications}}                                                             & \multirow{2}{*}{\textbf{\begin{tabular}[c]{@{}c@{}}Min. Required \\ Range\end{tabular}}} \\ \cline{1-3}
\multicolumn{1}{|c|}{\textbf{Cruise Speed}} & \multicolumn{1}{c|}{\textbf{Turn Rate}}   & \textbf{Vertical Speed} &                                                                                          \\ \hline
\multicolumn{1}{|c|}{30 kts}                & \multicolumn{1}{c|}{63.05 deg/s}          & 250-500 ft/min          & 1222m                                                                                    \\ \hline
\multicolumn{1}{|c|}{60 kts}                & \multicolumn{1}{c|}{10.51deg/s}           & 250-500 ft/min          & 963m                                                                                     \\ \hline
\multicolumn{1}{|c|}{\textbf{60 kts}}       & \multicolumn{1}{c|}{\textbf{31.53 deg/s}} & \textbf{250-500 ft/min} & \textbf{703m} ($\sim$ 700m)                                                              \\ \hline
\multicolumn{1}{|c|}{90 kts}                & \multicolumn{1}{c|}{7.01 deg/s}           & 250-500 ft/min          & 1018m                                                                                    \\ \hline
\multicolumn{1}{|c|}{\textbf{90 kts}}       & \multicolumn{1}{c|}{\textbf{21.02 deg/s}} & \textbf{250-500 ft/min} & \textbf{666m} ($\leq$ 700m)                                                              \\ \hline
\end{tabular}
\caption{\small Required min. detection range with P(track) $\geq$ 95\% based on maximum angular-rate error of 0.9deg/s for different UAS configurations. These values are taken from the ASTM F3442/F3442M standard.}
 \vspace{-1.0em}
\label{tab:ownship_classes} 
\end{table}
Based on Table~\ref{tab:ownship_classes} we can observe that our DAA surveillance system can satisfy low-risk UA operations for two specification categories: 
\begin{itemize}
    \item 60kts, 31.53deg/s, 250-500ft/min
    \item 90kts, 21.02deg/s, 250-500ft/min
\end{itemize}

The performance speed of the vision setup using NVIDIA Xavier AGX dev kit with and without TensorRT are 66.8 FPS and 5.3 FPS, respectively. 



\section{Conclusion}

We present, AirTrack, a state-of-the-art vision-based detection and tracking module for long-range aircraft detect-and-avoid applications. AirTrack uses cascaded detection modules along with a secondary classifier to improve performance. Comparative results on the Amazon AOT dataset as well as extensive real world flight tests showcase the efficacy of the proposed approach. We also interpret the results for the newly established ASTM standards and prove that AirTrack satisfies the standards for at least two specification categories. 

\section*{ACKNOWLEDGMENT}

The authors would like to thank Jan Hansen-Palmus, Joao Dantas, Ian Higgins, and, David Kohanbash for helping out with experiments, field tests, repair work, and discussions.


\bibliographystyle{ieeetr}
\bibliography{ref}

\end{document}